%% file: main.tex
\documentclass[10pt,twocolumn,letterpaper]{article}

\usepackage[pagenumbers]{cvpr} 

\input{preamble}

\definecolor{cvprblue}{rgb}{0.21,0.49,0.74}
\usepackage{multirow}
\usepackage{arydshln}
\usepackage[linesnumbered,ruled,vlined]{algorithm2e}
\usepackage[pagebackref,breaklinks,colorlinks,citecolor=cvprblue]{hyperref}

\def\confName{CVPR}
\def\confYear{2024}

\title{CR-SFP: Learning Consistent Representation for Soft Filter Pruning}

\author{
Jingyang Xiang$^1$
\and
Zhuangzhi Chen$^2$
\and
Jianbiao Mei$^1$
\and
Siqi Li$^1$
\and
Jun Chen$^1$
\and
Yong Liu$^1$\thanks{Corresponding author}
\and
\centerline{$^1$APRIL Lab, Zhejiang University, Hangzhou, China}
\and
\centerline{$^2$IVSN, Zhejiang University of Technology, Hangzhou, China}
}

\begin{document}
\maketitle
\input{sec/0_abstract}    
\input{sec/1_intro}
\input{sec/2_related_work}
\input{sec/3_method}

\input{sec/4_exp}

\input{sec/5_conclusion}

{
    \small
    \bibliographystyle{ieeenat_fullname}
    \bibliography{main}
}


\end{document}

%% file: preamble.tex
\usepackage[dvipsnames]{xcolor}


%% file: sec/0_abstract.tex
\begin{abstract}
Soft filter pruning~(SFP) has emerged as an effective pruning technique for allowing pruned filters to update and the opportunity for them to regrow to the network.
However, this pruning strategy applies training and pruning in an alternative manner, 
which inevitably causes inconsistent representations between the reconstructed network~(R-NN) at the training and the pruned network~(P-NN) at the inference, resulting in performance degradation.
In this paper, we propose to mitigate this gap by learning consistent representation for soft filter pruning, dubbed as CR-SFP.
Specifically, for each training step, CR-SFP optimizes the R-NN and P-NN simultaneously with different distorted versions of the same training data, 
while forcing them to be consistent by minimizing their posterior distribution via the bidirectional KL-divergence loss.
Meanwhile, the R-NN and P-NN share backbone parameters thus only additional classifier parameters are introduced.
After training, we can export the P-NN for inference.
CR-SFP is a simple yet effective training framework to improve the accuracy of P-NN without introducing any additional inference cost.
It can also be combined with a variety of pruning criteria and loss functions.
Extensive experiments demonstrate our CR-SFP achieves consistent improvements across various CNN architectures.
Notably, on ImageNet, our CR-SFP reduces more than 41.8\% FLOPs on ResNet18 with 69.2\% top-1 accuracy, improving SFP by 2.1\% under the same training settings.
%
The code will be publicly available on GitHub.
\end{abstract}

%% file: sec/1_intro.tex
\section{Introduction}
\label{sec:intro}
In recent years, deep neural networks~(DNNs) have achieved remarkable success in a variety of pattern recognition tasks, including image classification~\cite{simonyan2015very,he2016deep}, object detection~\cite{he2017mask, girshick2015fast}, semantic segmentation~\cite{girshick2014rich, croitoru2019unsupervised} and other fields.
However, the great success of DNNs often relies on huge parameters and floating point operations.
It hinders them from running in real-time in edge scenarios where computing power and memory are limited, such as smartphones, watches and other wearable devices.
In order to balance the trade-off between performance and inference cost, researchers have conducted extensive research on DNNs compression, including low-rank decomposition~\cite{denton2014exploiting}, model pruning~\cite{lin2020hrank}, model quantization~\cite{han2015learning}, knowledge distillation~\cite{hinton2015distilling} and efficient architecture design~\cite{zhang2018shufflenet, ding2021repvgg}.

\begin{figure}[t]
    \centering
    \includegraphics[width=0.45\textwidth]{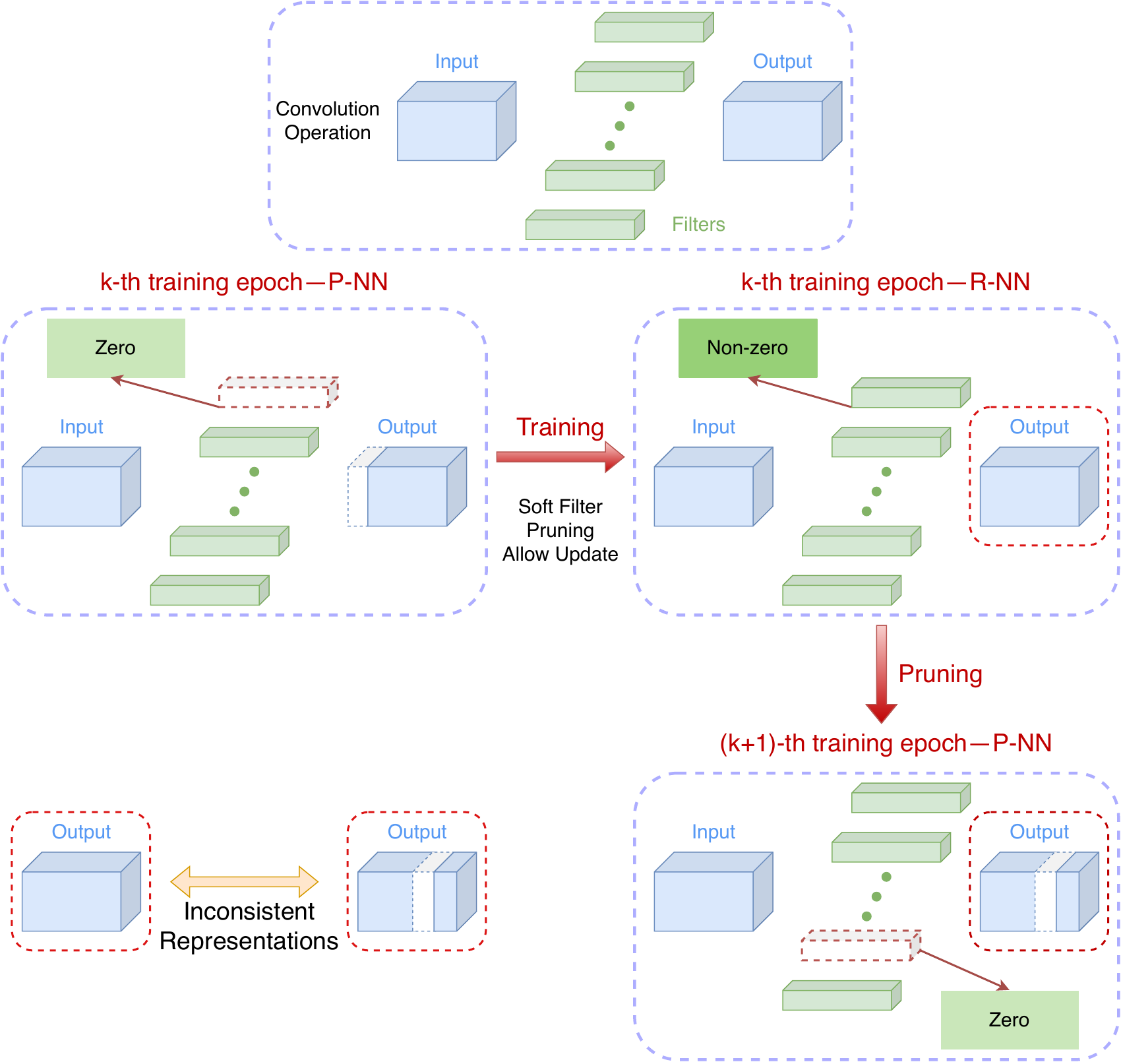}
    \vspace{-0.1in}
    \caption{\textbf{Inconsistent Representations in the Soft Filter Pruning}. We mark the pruned filters as the red dashed box. SFP allows the pruned filters to be updated during the training process so that the capacity of the model can be recovered. However, SFP trains and prunes network alternately, leading to inconsistent representations in the training and inference stage and hurting the final performance of the model.}
    \vspace{-0.2in}
    \label{inconsistent}
\end{figure}

\begin{figure*}[t]
\centering
\includegraphics[width=\textwidth]{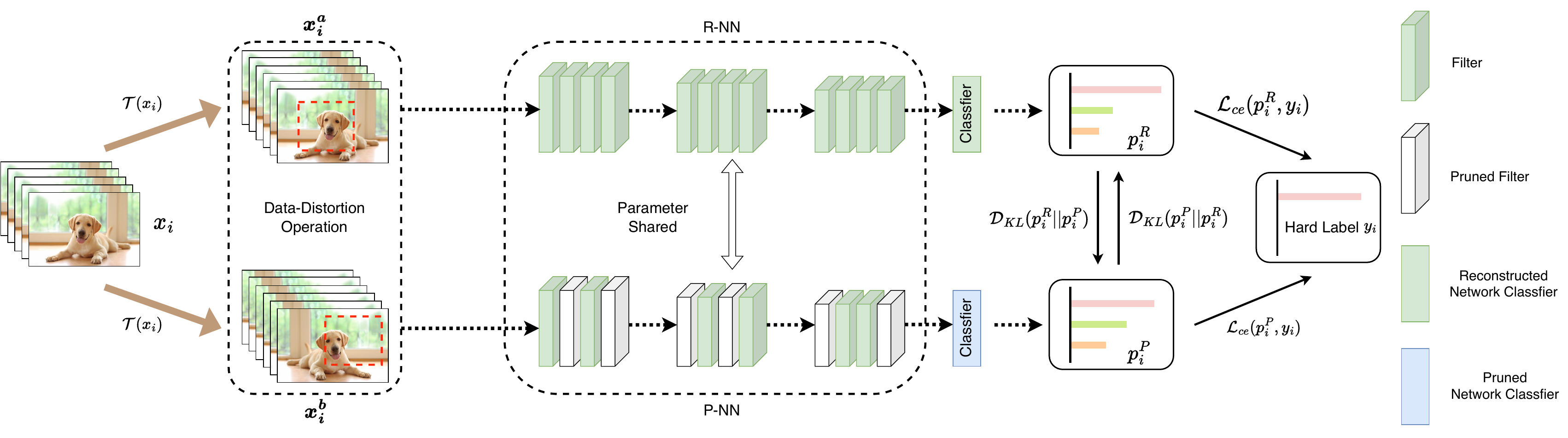} 
\vspace{-0.1in}
\caption{\textbf{The overall framework of our proposed CR-SFP}. We take image classfication task for illustration. The same input data with different data distortion operations is fed into R-NN and P-NN respetively. R-NN and P-NN share parts of parameters in the backbone and use different classifiers. Both of them are trained with hard label while reducing the representation gap between each other.}
\label{pipeline}
\vspace{-0.1in}
\end{figure*}

%
Among them, model pruning, which reduces memory and computation burden by removing unimportant weights or filters from the model, has received much attention from both industry and academia.
In the different prune type, most researchers have paid their attention to filter pruning.
Unlike weight pruning, which may still be less efficient in saving inference latency and require dedicated software and hardware to achieve actual acceleration,
filter pruning removes the entire filter so that it is hardware-friendly and well-compatible with Basic Linear Algebra Subprograms~(BLAS) libraries on general-purpose hardware.
Therefore, it is easy to achieve realistic acceleration.
Filter pruning has been extensively studied, and the current research on filter pruning can mainly be categorized into filter importance measurement~\cite{Dmolchanov2016pruning, he2019filter, lin2020hrank}, pruning process~\cite{shen2022prune}, pruning way~\cite{he2018soft, hou2022chex}.

In this paper, we focus on the pruning way.
According to whether the removed filters can regrow to the network during the training process, the pruning way can be divided into hard filter pruning~(HFP) and soft filter pruning~(SFP)~\cite{he2018soft}.
For HFP, the pruned filters are permanently removed during the training process. Thus the capacity of the model is reduced, which significantly harms the model's performance, particularly as the fraction of the network being pruned increases~\cite{humble2022soft}.
Most pretrain-prune-finetune~(PPF) pipeline for pruning belongs to HFP.
HRank~\cite{lin2020hrank} is a representative method, which prunes the filter that generates the low rank features and needs hundreds of epochs to finetune.
On the contrary, SFP allows the removed filters to be updated with gradients and regrow to the network during the training, preserving the model's capacity and achieving better performance.
Most current SOTA pruning methods are based on the SFP.
However, as shown in \cref{inconsistent}, the conventional SFP~\cite{he2018soft} follows the training-pruning pipeline alternately, resulting in inconsistent representations between the reconstructed network~(R-NN) during the training and the pruned network~(P-NN) during the inference, which hurts final performance of the P-NN.

To address the above-mentioned problem, we propose a novel approach to learn consistent representation for soft filter pruning~(CR-SFP).
The CR-SFP dynamically prunes the filters in a soft manner while narrowing the gap between P-NN and R-NN to maintain a consistent representation.
Particularly, before each training epoch, the filters among all network layers with minor importance will be selected and set to zero.
Then R-NN and P-NN will be trained simultaneously in a parameter-shared manner.
This approach avoids introducing a large number of additional training parameters.
At the same time, CR-SFP uses bidirectional KL-divergence loss to regularize posterior distribution~(class probabilities) in the update.
For a mini-batch input, we utilize the same data-distortion operation two twice to obtain two-branch distorted samples to train them respectively.
It prevents them overfitting the training set and be less effective due to the limitation of valuable information beyond the common hard label~\cite{anil2018large},
while boosting the generalization ability of model~\cite{simard2002transformation}.
Before the next training epoch, CR-SFP follows the pruning process of the original SFP, where a new set of filters of minor importance is pruned and a part of filters will regrow to network.
This pipeline will be alternated during the training process until the network converges. 
After the last epoch, we export the P-NN and obtain the compact model.
The CR-SFP inherits the advantage that SFP can enable the P-NN to have a larger model capacity.
Meanwhile, it optimizes R-NN and P-NN simultaneously while achieving consistent representation, thus achieving better performance.

Our main contributions are summarized as follows:
\begin{itemize}
\item To address the phenomenon of inconsistent representations of R-NN and P-NN in the SFP during training and inference, we propose to use bidirectional KL-divergence loss to regularize the posterior distribution and achieve consistent representation between them.
\item We use different data distortion operations and classifier layers for R-NN and P-NN to enhance the model's generalization capability and optimize them based on a deep mutual learning pipeline. R-NN and P-NN share a part of the backbone parameters. We only introduce a small number of additional parameters from the classfier layer and keep the training process simple and efficient.
\item We have conducted extensive experiments on ImageNet datasets to validate the effectiveness of our method. For example, our CR-SFP achieves 69.2\% top-1 accuracy for ResNet18 with 1.04G FLOPs, suppressing its counterpart SFP by 2.1\% under the same training settings.
\end{itemize}

%% file: sec/2_related_work.tex
\section{Related Work}
\label{sec:related_work}
\subsection{Network Pruning}
Network pruning aims to remove unimportant or redundant parameters in the complex model while preserving the origin performance.
In addition to the categorization based on the pruning way described above, the mainstream pruning type can be divided into weight pruning~\cite{han2015learning}, filter pruning~\cite{lin2020hrank, humble2022soft}, weight block pruning~\cite{elsen2020fast, tan2022accelerating, 2023chenrgp,1xn2023lin} and N:M pruning~\cite{mishra2021accelerating, pool2021channel} according to their pruning granularity.
Filter pruning allows for efficient speedup on general-purpose hardware.
The weight pruning is less efficient in memory access and computation, making it difficult to achieve realistic acceleration.
Weight block pruning and N:M pruning are semi-structured sparse models proposed in the recent years, both of which can rely on specific hardware or software for efficient acceleration.
They are regarded as a promising approach to achieve better trade-offs in inference latency and accuracy compared to weight pruning or filter pruning.
On the other hand, identifying important weights/filters/blocks is also an important aspect.
Although most of the work uses filter norm, absolute magnitude of weight, or scaling factors in batch normalization as pruning criteria, there are also some studies on importance measurement.
For example, FPGM~\cite{he2019filter} proposes to use the geometric median of filters to estimate the importance of filters.
The first-order Taylor expansion~\cite{Dmolchanov2016pruning, molchanov2019importance}, second-order derivatives~\cite{dong2017learning, peng2019collaborative, zheng2022savit} have also been proposed to approximate the loss change after pruning.
In order to obtain pruned networks more efficiently, researchers also investigate progressive pruning~\cite{hou2022chex}, pruning at initialized~\cite{hoang2022revisiting} to replace the complicate pretrain-pruning-finetune~\cite{lin2020hrank} pruning pipeline and these methods also obtain good results.

\subsection{Consistent Training}
Our work aims to learn consistent representation between R-NN and P-NN, which also relates to a few works of consistent training.
From a structure perspective, the most representative methods are ELD~\cite{ma2017dropout}, FD~\cite{zolna2018fraternal} and R-Drop~\cite{wu2021r}.
Specifically, the purpose of ELD is to reduce the gap between the dropout-enabled sub-model during training and the full model without dropout during inference, while FD works to enhance the representation consistency between the two sub-models.
Both ELD and FD adopt the $\ell_2$-norm to align the hidden space.
Different from them, R-Drop utilizes bidirectional KL-divergence loss to regularize the posterior distribution between the two sub-models, which improves the consistency of representation more effectively, and it also proves this constraint can improve the representation consistency of the model during the training and inference.
From a data perspective, DNNs with good generalization ability should exhibit similar posterior distributions for different data distortion operations of the same input data~\cite{simard2002transformation}.
DDGSD~\cite{xu2019data} excavates the potential capacity of a single network by learning consistent global feature distributions and posterior distributions across these distorted versions of data.
Cutoff~\cite{shen2020simple} discards part of the information within an input sentence to yield its restricted views and use them to enhance the consistency training.
To learn more consistent representation from data, researchers also have designed some new loss terms to enforce the features toward intra-class compactness and inter-class separability, 
such as contrastive loss~\cite{hadsell2006dimensionality}, triplet loss~\cite{schroff2015facenet} and center loss~\cite{wen2016discriminative}.

\subsection{Knowledge Distillation}
Minimizing the KL-divergence between the posterior distributions of teacher and student models is closely associated with knowledge distillation~\cite{hinton2015distilling, furlanello2018born, allen2023towards}.
Knowledge distillation improves the final performance of the student network by pre-training the teacher and then guiding the student or co-training the teacher and student.
In our setting, the teacher and student are the R-NN and P-NN, respectively.
R-NN and P-NN share a part of backbone weights in the network and use different classifier layers for themselves. Thus it resembles both self-knowledge distillation~\cite{xu2019data, mobahi2020self} and online knowledge distillation~\cite{zhang2018deep} scenario.
However, different from previous work, which aimed at improving the performance of the dense network by exploring dark knowledge from itself or distilling knowledge between layers~\cite{gotmare2018a, zhang2019your}, our work focuses on obtaining better P-NN from R-NN by learning consistent representation between them.
Moreover, unlike most previous methods, which introduce auxiliary networks and increase the complexity of the training process in need of large memory and time cost, 
our method only requires a single network via a parameter-shared approach, which is simple and efficient.

%% file: sec/3_method.tex
\section{Methodology}
In this section, we will firstly introduce the symbol and annotations to define a convolutional neural network~(CNN) and training procedure to SFP.
Then we will describe consistent learning process to our CR-SFP.
Finally, we will show how to obtain compact model.
\subsection{Preliminaries}
%
%
Without loss of generality, we consider a DNN with $L$ CNN layers.
We parameterize the weights in the CNN as $\mathbf{W}=\{\mathbf{W}^{1},...,\mathbf{W}^{L}\}$, 
where $\mathbf{W}^{l} \in \mathbb{R}^{C_{out}^{l} \times C_{in}^{l} \times K_h^{l} \times K_w^{l}}$ is the weight for $l$-th layer of the network.
$C_{out}^{l}$, $C_{in}^{l}$, $K_h^{l}$ and $K_w^{l}$ denote the output channel, input channel, kernel height and kernel width of the $l$-th layer respectively.
The shapes of input tensor $\mathbf{X}^l$ and output tensor $\mathbf{X}^{l+1}$ are $C_{in}^l \times H_l \times W_l$ and $C_{out}^l \times H_{l+1} \times W_{l+1}$.
The convolutional operation of the $j$-th filter in the $l$-th layer can be written as:
\begin{equation}
\label{stand_conv}
\mathbf{X}^{l+1}_j=\mathbf{W}_j^l \ast \mathbf{X}^l
\end{equation}
where $\mathbf{W}_j^l\in \mathbb{R}^{C_{in}^{l} \times K_h^{l} \times K_w^{l}}$ represents the $j$-th filter of $l$-th CNN layer.
and $\mathbf{W}^l$ consists of $\{\mathbf{W}_1^l,...,\mathbf{W}_{C_{out}^{l}}^l\}$.
The $\mathbf{X}^{l+1}_j$ represents the $j$-th output feature map of the $l$-th layer.
For simplify, we omit the normalization and activation layers of the network.
In the \cref{sec:compact_model}, we will describe how to deal with them to obtain compact model.

In order to obtain P-NN during the training, we introduce an output filter mask $\mathbf{m}^l\in\{0,1\}^{C_{out}^l}$ which is updated according to the pruning action for layer $l$.
Using these masks, we define the forward propagation of P-NN as:
\begin{equation}
\mathbf{X}^{l+1}_j=(\mathbf{W}_j^l \odot \mathbf{m}_j^l)  \ast \mathbf{X}^l=\hat{\mathbf{W}}_j^l \ast \mathbf{X}
\end{equation}
where $\mathbf{m}_j^l$ denotes whether the $j$-th filter is pruned.
$\mathbf{m}_j^l$ will be broadcasted to match the shape of $\mathbf{W}_j^l$. 

SFP divides the pruning pipeline into three steps: filter selection, pruning and reconstruction.
Specifically, during the filter selection stage, SFP evaluates the importance of each filter based on specific criteria. 
Assume that the pruning rate of the $l$-th layer is $P^l$.
In the filter pruning stage, we select the lowest important $C_{out}^{l}P^l$ filters by $\mathcal{I}(\mathbf{W}^l)$ and set them to zero to temporarily eliminate their contribution to the network output.
Unlike HFP, which permanently removes the pruned filters, SFP allows the pruned filters to be reconstructed (\eg updated from zero) by the forward-backward process and regrow to the network.
The SFP repeats these steps alternately during the whole training process, dynamically removes the filters and finally exports a compact and efficient model.

In this paper we use the $\ell_2$-norm to measure the importance of each filter:
\begin{equation}
\label{importance}
\mathcal{I}(\mathbf{W}_j^l)=
\sqrt[2]{\sum_{n=1}^{C_{in}^l}~\sum_{k_1=1}^{K_h^{l}}~\sum_{k_2=1}^{K_h^{l}}
\left | \mathbf{W}_j^l(n,k_1,k_2)\right |^2} 
\end{equation}
%
Meanwhile, we use the same pruning rate $P^l=P$  for each layer to reduce hyper-parameters.
It's worth noting that our approach applies equally to other criteria or cases where different layers have different pruning rates.

\subsection{Consistent Representation}
We define the training set $\mathcal{S}=\{(x_1,y_1),...,(x_N,y_N)\}$ as a labeled source dataset, 
where $N$ is the total number of training samples and $y_i$ is the corresponding hard label.
Then the optimization objection for P-NN $\phi\left({x}_{i} ; \mathbf{W}, \mathbf{m}\right)$ can be formulated as:
\begin{equation}
\underset{\mathbf{W}, \mathbf{m}}{\operatorname{argmin}} \frac{1}{N} \sum_{i=1}^{N} \mathcal{L}_{ce}\left(\phi\left({x}_{i} ; \mathbf{W}, \mathbf{m}\right), y_i\right)
\label{eq:problem}
\end{equation}
where $\mathcal{L}_{ce}$ is the cross-entropy loss.

Conventional SFP~\cite{he2018soft} optimizes \cref{eq:problem} via minimizing the loss for R-NN $\phi\left({x}_{i} ; \mathbf{W}, \mathbf{1}\right)$ via:
\begin{equation}
\underset{\mathbf{W}, \mathbf{m}}{\operatorname{argmin}} \frac{1}{N} \sum_{i=1}^{N} \mathcal{L}_{ce}\left(\phi\left({x}_{i} ; \mathbf{W}, \mathbf{1}\right), y_i\right)
\label{eq:rnn}
\end{equation}
which neglects the representation gap between P-NN and R-NN and inevitably hurts the performance to P-NN.
%

%
To address the above-mentioned problem, we propose CR-SFP in this paper, a simple but effective framework to learning consistent representation for P-NN and R-NN.
The overall framework of our method is shown in \cref{pipeline}.
Specifically, for the one training step, we sample a mini-batch input $B=\{(x_1,y_1), ..., (x_n,y_n)\}$ from training set $S$ and apply the same data-distortion operation $\mathcal{T}(B)$ two twice to obtation two-branch distorted samples
$B^{a}=\{(\boldsymbol{x_1^a},y_1), ..., (\boldsymbol{x_n^a},y_n)\}$ and $B^{b}=\{(\boldsymbol{x_1^b},y_1), ..., (\boldsymbol{x_n^b},y_n)\}$, where the hard labels $\{y_1,...,y_n\}$ are shared for them.
Then, $B^a$ and $B^b$ are fed into R-NN and P-NN to get corresponding outputs and their cross entropy loss:
\begin{equation}
\left\{\begin{split}
 p_i^R &= \mathop{softmax}\left(\phi\left(\boldsymbol{{x}_{i}^a} ; \mathbf{W}, \mathbf{m}\right)\right),i=1,...,n \\
 p_i^P &= \mathop{softmax}\left(\phi\left(\boldsymbol{{x}_{i}^b} ; \mathbf{W}, \mathbf{1}\right)\right),i=1,...,n \\
 \mathcal{L}_{ce}&=\frac{1}{n}\sum_{i=1}^{n}\mathcal{L}_{ce}(p_i^R, y_i)+\mathcal{L}_{ce}(p_i^P, y_i)
\end{split}\right.
\label{eq:compute_threshold}
\end{equation}

%

%
To reach the consistent posterior representation between the two-branch distorted versions, 
following the previous works~\cite{zhang2018deep, wu2021r}, CR-SFP adopts the bi-directional KL divergence to measure the match as follows:
\begin{equation}
\mathcal{L}_{KL}=\frac{1}{2n}\sum_{i=1}^{n}\mathcal{D}_{KL}\left({{p}_i^R}|| {{p}_i^P} \right )+\mathcal{D}_{KL}\left({{p}_i^P}||{{p}_i^R} \right )
\label{bi_kdloss}
\end{equation}
where
\begin{equation}
    \begin{aligned}
        \mathcal{D}_{KL}\left( p_1||p_2\right )=-\sum p_1\operatorname{log}p_2+\sum p_1\operatorname{log}p_1
    \end{aligned}
    \label{kdloss}
\end{equation}

%
Similar to \cite{zhang2018deep, chen2021exploring}, we implement $\operatorname{stopgrad}$ operation by modifying \cref{kdloss} as:
\begin{equation}
\mathcal{D}_{KL}\left(\operatorname{stopgrad} \left (p_1\right )||p_2\right )
\label{stopgrad}
\end{equation}
This means that $p_1$ is treated as a constant in this term and the gradient to $\partial D_{KL}\left(p_1||p_2\right)/ \partial p_1$ is zero.

Our CR-SFP mainly consists of two types of loss terms to update parameters of single network:
\begin{equation}
\mathcal{L}_{net}=\mathcal{L}_{ce}+\lambda \mathcal{L}_{KL}
\label{final_loss}
\end{equation}
where $\lambda$ is bi-directional KL-divergence loss weight for posterior distribution regularization.
The first term of \cref{final_loss} are the supervised losses of R-NN and P-NN. We improve their performance by simultaneously optimizing the corresponding parameters of both networks in a paramater-shared manner.
The last term aims to improve the consistency of the R-NN and P-NN representation and further enhance the performance to P-NN.


\begin{table*}[t]
\centering
\normalsize
\setlength{\tabcolsep}{4mm}
\begin{tabular}{@{}llcccccc@{}}
\toprule
\textbf{Model}             & \textbf{Method} & \textbf{PT} & \textbf{FLOPs} & \textbf{Parameters} & \multicolumn{1}{l}{\textbf{Top-1}} & \multicolumn{1}{l}{\textbf{Top-5}} & \textbf{Epochs} \\ \midrule
\multirow{5}{*}{ResNet18}  & MIL~\cite{dong2017more}                & N        & 1.17G   & N/A       & 66.3\%                                 & 86.9\%                                 & 200             \\
                           & SFP~\cite{he2018soft}                  & N        & 1.04G   & 6.53M     & 67.1\%                                 & 87.8\%                                 & 100             \\
                           & FPGM~\cite{he2019filter}               & N        & 1.04G   & 6.53M     & 67.8\%                                 & 88.1\%                                 & 100             \\ 
                           & PFP~\cite{liebenwein2019provable}      & Y        & 1.27G   & 6.57M     & 67.4\%                                 & 87.9\%                                 & 270             \\
                           & \textbf{CR-SFP~(Ours)}   & \textbf{N}        & \textbf{1.04G} & \textbf{6.53M}         & \textbf{69.2\%}         & \textbf{88.9\%}           & \textbf{100}             \\ 
\midrule
\multirow{9}{*}{ResNet34}  & SFP~\cite{he2018soft}                  & N        & 2.2G & 11.10M       & 71.8\%                                 & 90.3\%                                 & 100             \\ 
                           & FPGM~\cite{he2019filter}               & Y        & 2.2G & 11.10M       & 72.6\%                                 & 91.1\%                                 & 200             \\
                           & DMC \cite{gao2020discrete}             & Y        & 2.1G & N/A          & 72.6\%                                 & 91.1\%                                 & 490             \\
                           & SCOP~\cite{tang2020scop}               & Y        & 2.0G & 11.86M       & 72.6\%                                 & 91.0\%                                 & 230             \\
                           & CHEX~\cite{hou2022chex}                & N        & 2.0G & 15.43M       & 73.5\%                                 & N/A                                    & 250             \\
                           & NPPM~\cite{gao2021network}             & Y        & 2.1G & N/A          & 73.0\%                                 & N/A                                    & 390             \\
                           & W-Gates~\cite{li2022weight}            & Y        & 2.1G & N/A          & 72.7\%                                 & N/A                                    & 210             \\
                           & HTP-URC~\cite{qian2023htp}             & Y        & 2.1G & 12.00M       & 73.0\%                                 & N/A                                    & 200             \\ 
                           & \textbf{CR-SFP~(Ours)}  & \textbf{N}   & \textbf{2.2G}   & \textbf{11.10M}      & \textbf{73.0\%}                     & \textbf{91.1\%}            & \textbf{100}  \\
\midrule
\multirow{10}{*}{ResNet50} & SFP~\cite{he2018soft}                  & N        & 2.3G  & 15.93M      & 74.6\%                                 & 92.9\%                                 & 100             \\
                           & FPGM~\cite{he2019filter}               & Y        & 2.3G  & 15.93M      & 75.6\%                                 & 92.9\%                                 & 200             \\
                           & SCOP~\cite{tang2020scop}               & Y        & 2.2G  & 14.62M      & 76.0\%                                 & 92.8\%                                 & 230             \\
                           & EagleEye~\cite{li2020eagleeye}         & Y        & 2.0G  & N/A         & 76.4\%                                 & 92.9\%                                 & 240             \\
                           & Taylor~\cite{molchanov2019importance}  & Y        & 2.3G  & 14.20M      & 74.5\%                                 & N/A                                    & 125             \\
                           & DSA~\cite{ning2020dsa}                 & N        & 2.0G  & N/A         & 74.7\%                                 & 92.1\%                                 & 120             \\
                           & While-Box~\cite{zhang2022carrying}     & N        & 2.2G  & N/A         & 75.3\%                                 & 92.4\%                                 & 190             \\
                           & W-Gates~\cite{li2022weight}            & Y        & 2.3G  & N/A         & 75.7\%                                 & 92.6\%                                 & 210             \\
                           & TPP~\cite{wang2023trainability}        & Y        & 2.5G  & N/A         & 76.4\%                                 & N/A                                    & 190             \\
                           & DepGraph~\cite{fang2023depgraph}       & Y        & 2.0G  & 12.46M      & 75.8\%                                 & N/A                                    & 190             \\
                           & \textbf{CR-SFP~(Ours)} & \textbf{N}        & \textbf{2.3G}  & \textbf{15.93M}         & \textbf{76.6\%}& \textbf{93.1\%}              & \textbf{100}             \\
                           
\bottomrule
\end{tabular}
\caption{\textbf{Comparison of pruning ResNet on ImageNet}. In ``FT'' column, ``Y'' and ``N'' indicate whether requiring the pre-trained model as initialization or not. We report FLOPs, Parameters, top-1 and top-5 accuracy of pruned model. ``Epoch'' are reported as: pretraining epochs~(if needed) plus all subsequent training epochs to obtain the final pruned model.}
\label{tab:main_result}
\vspace{-0.1in}
\end{table*}

\subsection{Obtaining Compact Model}
\label{sec:compact_model}
\begin{figure}[ht]
    \centering
    \includegraphics[width=0.48\textwidth]{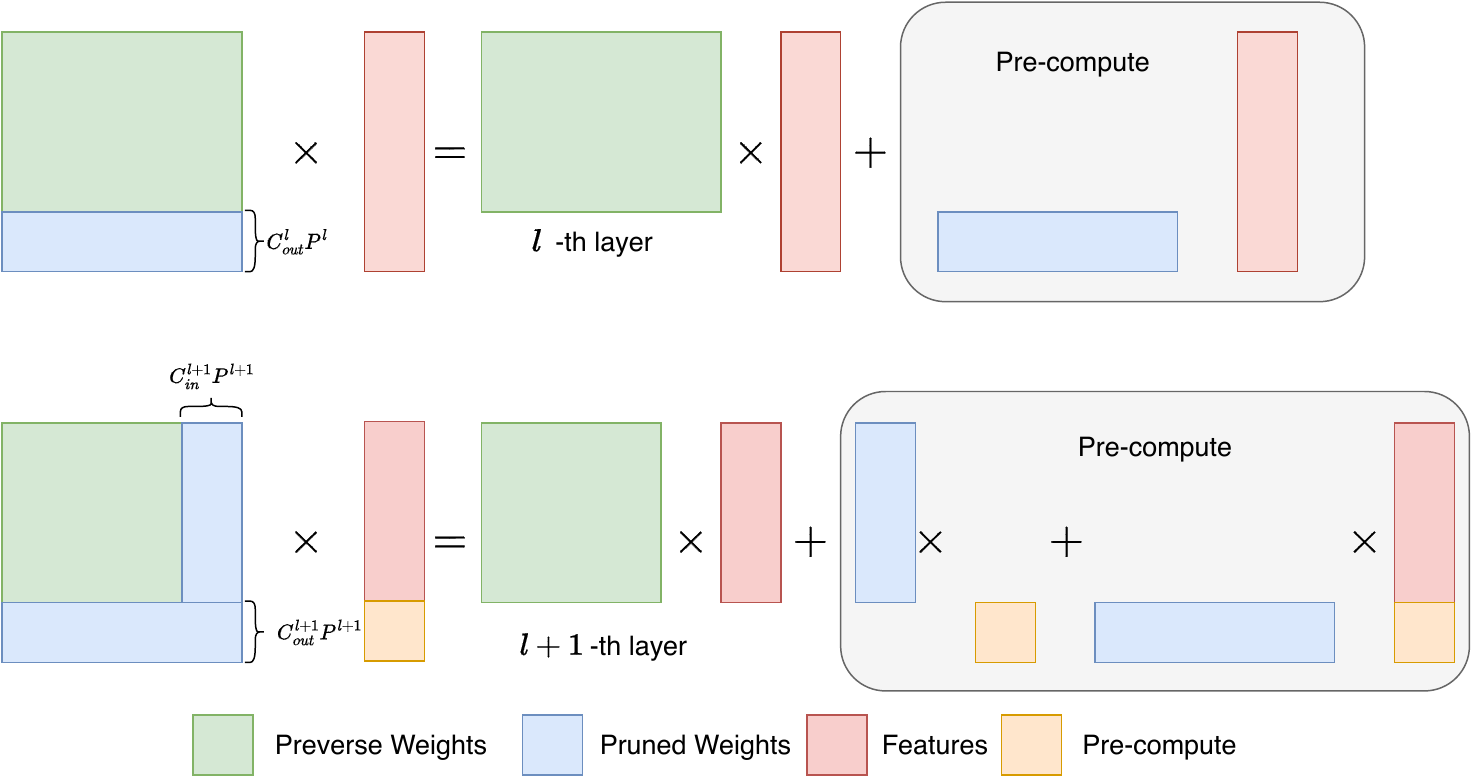}
    \caption{\textbf{Pipeline for obtaining compact model}. We ommit the BN and act layer for simplify.}
    \vspace{-0.1in}
    \label{fig:pipline_compact}
\end{figure}
%
Pruning filters is equivalent to removing corresponding the output feature maps in the $l$-th layer.
Although the presence of the normalization and activation layer between the $l$-th and the $l+1$-th convolutional layer may result in a non-zero feature map,
these values can be pre-computed and merged into the $l+1$-th normalization layer during the inference process without introducing additional overhead.
Thus, as the output channel of the $l$-th convolutional layer is reduced from $C_{out}^l$ to $C_{out}^l(1-P^l)$, 
the input channel of the $l+1$-th convolutional layer will also be reduced from $C_{in}^{l+1}$ to $C_{in}^{l+1}(1-P^{l})$.
In this case, the computational cost of the model will be significantly reduced and we can achieve an efficient acceleration ratio.

Here we give specific steps for obtaining compact model as shown in \cref{fig:pipline_compact}.
Assuming that we complete the pruning of  $l$-th convolutional layer, 
we can split its weights $\mathbf{W}^l$ into the pruned weights $\mathbf{W}^l_{p}$ and the retained weights $\mathbf{W}^l_{r}$, 
where $\mathbf{W}^l=\left[\mathbf{W}^l_{p};\mathbf{W}^l_{r}\right]$. 
Based on \cite{pool2021channel}, we can know our description is not out of generality.
The forward propagation of the $l$-th layer can be written as:
\begin{equation}
\begin{aligned}
\mathbf{X}^{l+1}=\left[\mathbf{X}^{l+1}_{p};\mathbf{X}^{l+1}_{r}\right]
=\left[\mathbf{W}^l_{p} \ast \mathbf{\hat{X}}^l;\mathbf{W}^l_{r} \ast \mathbf{\hat{X}}^l\right ]
\end{aligned}
\label{eq::begin_1}
\end{equation}
where $\mathbf{X}^{l+1}_{p}$ and $\mathbf{X}^{l+1}_{r}$ are the corresponding features for $\mathbf{W}^l_{p}$ and $\mathbf{W}^l_{r}$ respectively.

We assume that the convolutional layer is followed by the batch normalization~(BN) and activation~(act) layer, then
$\mathbf{\hat{X}}^{k}=\operatorname{act}\left(\operatorname{BN}\left (\mathbf{{X}}^{k}\right) \right )$.
The forward propagation of the $l$-th layer can be written as:
\begin{equation}\begin{aligned}
\label{eq::begin}
\mathbf{\hat{X}}^{l+1} & = \operatorname{act}\left(\operatorname{BN}\left( \mathbf{X}^{l+1}\right) \right) 
\\ &= \left[\operatorname{act}\left(\operatorname{BN}\left( \mathbf{X}^{l+1}_{p}\right) \right);\operatorname{act}\left(\operatorname{BN}\left( \mathbf{X}^{l+1}_{r}\right) \right)\right]
\\ &= \left[\mathbf{\hat{X}}^{l+1}_{p};\mathbf{\hat{X}}^{l+1}_{r}\right]
\end{aligned}\end{equation}

Since we prune the weights $\mathbf{W}^l_{p}$, $\mathbf{X}^{l+1}_{p}$ is equal to zero and the parameters in BN and act are known ahead of time,
we can pre-compute $\mathbf{\hat{X}}^{l+1}_{p}$ in the process of inference.

For the forward propagation in the $l+1$-th layer:
\begin{equation}
\begin{aligned}
&\mathbf{X}^{l+2}=\left[\mathbf{X}^{l+2}_{p};\mathbf{X}^{l+2}_{r}\right]
=\left[\mathbf{W}^{l+1}_{p} \ast \mathbf{\hat{X}}^{l+1};\mathbf{W}^{l+1}_{r} \ast \mathbf{\hat{X}}^{l+1}\right ]
\\ &=\left[\mathbf{W}^{l+1}_{p} \ast \mathbf{\hat{X}}^{l+1};
\mathbf{W}^{l+1}_{r} \ast \left[\mathbf{\hat{X}}^{l+1}_{p};\mathbf{\hat{X}}^{l+1}_{r}\right]\right ]
\end{aligned}
\end{equation}
Based on the additivity of the convolution, we can get:
\begin{equation}
\begin{aligned}
& \mathbf{W}^{l+1}_{r} \ast \left[\mathbf{\hat{X}}^{l+1}_{p};\mathbf{\hat{X}}^{l+1}_{r}\right ] \\
& =\left[\mathbf{W}^{l+1}_{r_1};\mathbf{W}^{l+1}_{r_2}\right] \ast \left[\mathbf{\hat{X}}^{l+1}_{p};\mathbf{\hat{X}}^{l+1}_{r}\right ]\\
& = \mathbf{W}^{l+1}_{r_1} \ast \mathbf{\hat{X}}^{l+1}_{p}  + \mathbf{W}^{l+1}_{r_2} \ast \mathbf{\hat{X}}^{l+1}_{r}
\end{aligned}    
\label{input_prune}
\end{equation}
where $\mathbf{W}^{l+1}_{r_1}$ and $\mathbf{W}^{l+1}_{r_2}$ represent the corresponding indices of the $l+1$-th layer weight along the input channel based on the pruning results of the $l$-th layer.
Since $\mathbf{W}^{l+1}_{r_1} \ast \mathbf{\hat{X}}^{l+1}_{p}$ in the \cref{input_prune} can also be pre-computed and integrated into BN layer after the $l+1$-layer,
we can also remove the corresponding input channels and repeat this process to obtain a compact model.

The details of CR-SFP are explained in \cref{alg:cr-sfp}.

\setlength{\textfloatsep}{0.1cm}
\begin{algorithm}[t]
    \small
    \textbf{Input}: training data: $\mathcal{S}$, pruning rate: $P$, total training epochs $epoch_\text{max}$  \;
    \textbf{Parameter}: model parameters $\mathbf{W}=\{\mathbf{W}^{1},...,\mathbf{W}^{L}\}$, filter masks $\mathbf{m}=\{\mathbf{m}^1,...,\mathbf{m}^L\}$ \;
    \textbf{Output}: the pruned model and its parameters $\mathbf{W}^*$
    
    \For{$epoch=1;epoch\leq epoch_\text{max};epoch++$}{
         Update the model parameter $\mathbf{W}$ based on \cref{final_loss} \;
        \For{$l=1;l\leq L;l++$}{
            Get the $\mathcal{I}(\mathbf{W}_j^l)$ for each filter, $1\leq j \leq C_{out}^l$ \;
            Zeroize $C_{out}^l P$ filters by $\mathcal{I}(\mathbf{W}_j^l)$ filter selection \;
            Update  $\mathbf{m}^l$ based by $\mathcal{I}(\mathbf{W}_j^l)$ filter selection \;
        }
    }
    Reformat $\mathbf{W}$ to $\mathbf{W}^*$ based on \cref{eq::begin_1} to \cref{input_prune} \;
    \textbf{return} the pruned model and its parameters $\mathbf{W}^*$ \;
    \caption{Algorithm Description of CR-SFP}
    \label{alg:cr-sfp}
\end{algorithm}
\setlength{\floatsep}{0.1cm}

%% file: sec/4_exp.tex
\section{Experiments}
\subsection{Benchmark Datasets and Network Structure}
In order to investigate the effectiveness of our CR-SFP, we adopt ImageNet~\cite{russakovsky2015imagenet} as a benchmark dataset.
ImageNet is a large-scale dataset for image classification tasks, containing 1.28 million training images and 50,000 validation images with 1000 classes.
Following previous works, we focus on pruning the challenging ResNet models with different depths in this paper, including ResNet18, ResNet34, and ResNet50.
The baseline ResNet-18/34/50 models have 1.8/3.7/4.1 GFLOPs.

\textbf{Experimental Settings}.
Following the previous work SFP~\cite{he2018soft} and FPGM~\cite{he2019filter}, our network is trained from scratch rather than initialized on a pre-trained model.
We use stochastic gradient descent~(SGD) for all experiments with momentum of 0.9 and weight decay 1e-4 to optimize the training objective.
The initial learning rate is set to 0.1 for the total batch size of 256.
All networks are trained for 100 epochs and the learning rate is reduced by a multiple of 0.1 at 30,60 and 90 epochs respectively.
For data augmentation, we only apply random crops, horizontal flip, and normalization.
All experiments run on PyTorch~\cite{paszke2019pytorch} framework with NVIDIA RTX3090 GPUs using AMP~(automatic mixed precision).
We calculate FLOPs by counting multiplication and addition as one operation following He~\cite{he2016deep}.

To keep our method simple and generic, we set $\lambda=0.2$ to the bidirectional KL-divergence loss in Eq.\,(\ref{final_loss}) to all experiments unless otherwise stated.
At the same time, we did not try to adjust potential strategies such as the criterion of filter importance and the pruning interval epoch.
For different networks and datasets, their optimal values can be obtained using grid search with others fixed.
In this paper, though they are not optimal, we still achieved competitive results with current state-of-the-art methods.

\subsection{ResNet on ImageNet}
Same with SFP~\cite{he2018soft}, we prune ResNet models with the same pruning rate across layers and do not prune the shortcut branch for simplification.

\cref{tab:main_result} shows the experiment results for the ImageNet dataset.
CR-SFP shows excellent performance with ResNet models of different depths.
For ResNet18, CR-SFP can achieve 69.2\% top-1 accuracy with only 100 training epochs, exceeding SFP 2.1\%.
Moreover, it exceeds PFP by 1.8\% top-1 accuracy, despite PFP having 1.27G FLOPs and a total of 270 training epochs.
The results show that CR-SFP not only inherits the advantage that SFP can maintain the model's capacity but also improves the convergence speed by learning consistent representation of R-NN and P-NN, thus achieving better results with the same training epoch.
For ResNet34 and ResNet50, CR-SFP achieves competitive results with fewer epochs than the previous state-of-the-art methods.
Moreover, our method only needs to be trained from scratch, avoiding the complex process of training, pruning, and fine-tuning pipeline required by pre-training-based methods like \cite{li2020eagleeye, zhang2022carrying, li2022weight, qian2023htp}.

We do not adopt advanced training recipes on CR-SFP, such as cosine LR schedule, more extensive epochs or data augmentation such as mixup~\cite{zhang2018mixup} in \cref{tab:main_result}.
From \cref{sec:ablation}, we can see by applying these training recipes, our approach achieves better results. 


\subsection{Ablation Analysis}
\label{sec:ablation}
We use ablation analysis to investigate the effectiveness of different components in the CR-SFP.
All the following results are based on pruning the ResNet18 model to 1.04G FLOPs on the ImageNet.

\begin{table}[t]
\centering
\renewcommand\arraystretch{1.2}  
\begin{tabular}{@{}lcc@{}}
\toprule
\textbf{Method}                                                                                      & \textbf{Top-1} & \textbf{Top-5} \\ \midrule
\;$\mathcal{L}_{ce}(y_i,\hat{p}_i^R)$~(SFP Baseline)                                                  &  67.1\%         & 87.8\%         \\
\;\;+\;$\mathcal{L}_{ce}(y_i,\hat{p}_i^P)+0.2\mathcal{D}\left(\hat{p}_i^R,\hat{p}_i^P \right )$         &  68.6\%         &  88.3\%         \\
\;\;+\;Data Distortion~(\textbf{default})                                                            &  69.2\%         & 88.9\%          \\
\;\;+\;Advanced Training Recipe                                                                      &  69.5\%         & 89.1\%          \\ \bottomrule
\end{tabular}
\caption{Ablation study of different components in CR-SFP. Advanced training recipe means cosine LR schedule and longer training epochs~(200 epochs).}
\label{tab:component}
\end{table}

\begin{table}[t]
\centering
\begin{tabular}{@{}ccccc@{}}
\toprule
\multirow{2}{*}{\textbf{Method}} & \multicolumn{2}{c}{Baseline} & \multicolumn{2}{c}{Pruned} \\ \cmidrule(l){2-5} 
                       & Top-1  & Top-5  & Top-1  & Top-5  \\ \midrule
SFP (Baseline)         & 69.8\% & 89.1\% & 68.0\% & 88.1\% \\
\textbf{CR-SFP (Ours)} & 69.8\% & 89.1\% & 69.3\% & 88.8\% \\ \bottomrule
\end{tabular}
\caption{Results starting from the pretrained models.}
\label{tab:pretrained}
\end{table}

\textbf{Component Study.}
In \cref{tab:component}, we study the effectiveness of different components in CR-SFP.
The baseline is SFP~\cite{he2018soft}, where the model is pruned by minimize \cref{eq:rnn}.

Firstly, we apply the same data distortion operation for the input data with \cref{final_loss}), i.e., $\boldsymbol{x^a_i}=\boldsymbol{x^b_i}$.
This mutual learning process brings in 0.6\% top-1 accuracy for improvement.
When the different data distortion operations are applied, instead of sticking to the fixed input distributions during the whole training, we can obtain a P-NN with a stronger consistent representation with the R-NN.
It further improves the top-1 and top-5 accuracy by 0.3\% and 0.2\% respectively.
Following previous methods~\cite{su2021locally, hou2022chex} which applied more advanced training recipes for the pruning, we use a cosine LR schedule and longer training epochs~(200 epochs), prompting better model convergence.
Benefiting from the advanced training recipes, we achieve 69.5\% top-1 accuracy, further outperforming existing methods.

\textbf{CR-SFP from Pretrained Models.}
To further demonstrate the generality of our CR-SFP, we use the pre-trained ResNet18 weight to initialize the model.
We adopt the ResNet18 weights from torchvision~\cite{torchvision2016} model zoo for a fair comparison.
We perform 100 epochs of pruning on both SFP and CR-SFP.
As shown in \cref{tab:pretrained}, our method achieves 1.3\% top-1 accuracy improvement than the original SFP when under the same training settings and reducing the same amount of FLOPs.
What is more, comparing \cref{tab:component} and \cref{tab:pretrained}, we find that with the same training epoch~(200 versus 100+100), our method can obtain better results~(69.5\% versus 69.3\%) when trained from scratch than pretrain, prune and finetune pipeline.

\textbf{Pruning Criterion.}
To validate the effectiveness of CR-SFP, we apply this training framework to the different pruning criteria, 
including $\ell_2$-norm, Geometric Median~\cite{he2019filter}, Taylor FO~\cite{molchanov2019importance} 
and compare their final accuracy in \cref{tab:prune_criterion}.
We observe that CR-SFP is robust to the pruning criterion, as it improves the accuracy for all three criteria.
%
%
For example, with only 100 training epochs, CR-SFP with Geometric Median achieves 69.3\% top-1 accuracy, which further suppresses the results in \cref{tab:main_result}.
This suggests the learning of consistent representation and the study of prune criterion are vertical, and the two can be mutually enhanced to achieve better results.
\begin{table}[t]
\centering
\normalsize
\begin{tabular}{@{}ccccc@{}}
\toprule
\multirow{2}{*}{\textbf{Prune Criterion}} & \multicolumn{2}{c}{SFP (Baseline)} & \multicolumn{2}{c}{\textbf{CR-SFP (Ours)}} \\ 
\cmidrule(l){2-5} 
                 & Top-1 & Top-5 & Top-1 & Top-5 \\ \midrule
$\ell_2$-norm    & 67.1\%      & 87.8\%     & 69.2\%& 88.9\%\\ \midrule
Geometric Median & 67.8\%      & 88.1\%     & 69.3\%& 88.9\%\\ \midrule
Taylor FO        & 67.5\%      & 88.0\%     & 69.0\%& 88.8\%\\ \bottomrule
\end{tabular}
\caption{Comparison with different pruning criterion.}
\label{tab:prune_criterion}
\end{table}
\begin{table}[t]
\centering
\normalsize
\setlength{\tabcolsep}{15pt}
\begin{tabular}{@{}lcc@{}}
\toprule
\textbf{Distance Function}   & \textbf{Top-1}        & \textbf{Top-5}   \\ \midrule
{KL-Divergence}              & 69.2\%& 88.9\%\\ \midrule
{Cosine-Similarity}          & 68.9\%& 88.4\%\\ \midrule
{Binary-Crossentropy}        & 68.7\%& 88.5\%\\ 
\bottomrule
\end{tabular}

\caption{Comparison with different distance function.}
\label{tab:distance}
\end{table}

\textbf{Distance Function.}
We study the influence of distance function $\mathcal{D}$ to our CR-SFP, including bidirectional KL-divergence, negative cosine similarity and binary cross-entropy.
All hyper-parameters and architectures are unchanged and the operation of $\operatorname{stopgrad}$ is also applied as mentioned above, though they may be sub-optimal for this.

As can be seen in \cref{tab:distance}, our method works even better with KL-divergence, 
which bring 0.3\% and 0.5\% top-1 accuracy improvement respetively compared to negative cosine similarity and binary cross-entropy.
%
%
Since different distance functions should correspond using different hyper-parameters, we consider it is normal for cosine similarity and binary cross-entropy to outperform KL-divergence with default settings.
As previous distillation methods have often used KL-divergence as a loss function to regularize the posterior probability distribution, we adopt it as default in this paper.
We believe that better results can be achieved by choosing appropriate hyper-parameters based on specific distance functions.
%

\textbf{Effect of Weight $\lambda$.}
Further, we investigate the impact of the bidirectional KL-divergence loss weight $\lambda$. 
%
%
Here, we conduct experiments by varying the $\lambda$ in \cref{tab:lambda}.
For ResNet18, small $\lambda$ often brings better results and $\lambda=0.2$ is the best which achieves 69.2\% top-1 accuracy.
Too much regularization is not good because it will make network ignore information from labels and top-1 accuracy drops to 66.6\% when $\lambda$ is set to 5.
We also set $\lambda=-0.1.-0.2,-0.5$ respectively to enhance the inconsistency of R-NN and P-NN.
It only gets 65.5\% top-1 accuracy with $\alpha=-0.5$, which indicates the importance of consistent representation.
Although $\lambda=-0.1$ and $\lambda=-0.2$ also achieve good results, they are still down compared to $\lambda=0$.
We believe this is due to the fact that although these settings enhance inconsistency leading to performance degradation, they still achieve reasonable results due to strong monitoring of supervised losses $\mathcal{L}_{ce}$.
Note the choice of $\lambda$ should be related to the degree of model inconsistency and the model's ability to fit the data.
Therefore, adopting a more appropriate $\lambda$ for different models and datasets should yield better results.

\begin{table}[t]
\setlength{\tabcolsep}{3mm}
\centering
\normalsize
\begin{tabular}{@{}lcclcc@{}}
\toprule
$\boldsymbol{\lambda}$ & \bf Top-1 & \bf Top-5       & $\boldsymbol{\lambda}$ & \bf Top-1             & \bf Top-5             \\ \midrule
-0.5      & 65.5\%          & 86.4\%    & 0.2       & \bf 69.2\% & \bf 88.9\%\\
-0.2      & 68.3\%& 88.1\%& 0.5       & 68.9\%& 88.6\%\\
-0.1      & 68.5\%& 88.2\%& 1         & 68.9\%& 88.4\%\\
0         & 68.6\%& 88.3\%& 2         & 68.4\%& 88.2\%\\
0.1       & 68.8\%& 88.6\%& 5         & 66.6\%& 87.1\%\\ \bottomrule
\end{tabular}
\caption{Comparison with different weight $\lambda$.}
\label{tab:lambda}
\end{table}

\section{Cost Analysis}
Compared to the conventional SFP procedure, our implementation needs to forward batch data twice at each training step.
One potential limitation is that the computational increases at each step.
Hence, we plot the imagenet valididation accuracy curves along the training epoch of SFP, SFP with a doubled batch size and CR-SFP. 
We implement double batch size by applying input data $B$ with two twice data distorition operations and concatenating them to $\left[B^a;B^b\right]$ in the same mini-batch to forward once.
This is similar to enlarging the epochs to be double at training stage.
The difference is that half of the data in doubled batch size are the same as the other half, 
while directly doubling the training epoch, the data in the same mini-batch are all different.
The curves are shown in \cref{fig:convergence}.
We can see that double batch size can also achieve improvement to SFP~(68.4\% top-1 accuracy) while it still falls behind our CR-SFP~(69.2\% top-1 accuracy).
Meanwhile, we also test the training speed.
For the detailed training speed for each step, we present the number here:
SFP + Double Batch costs near 214ms per step, CR-SFP costs about 218ms per step.
The additional time cost comes from the KL-divergence loss backward computation, which is only 4ms and almost negligible.
From the above analysis, we can know CR-SFP is an efficient framework and improves the performance to a stronger one with similar training costs.

%% file: sec/5_conclusion.tex
\begin{figure}[t]
    \centering
    \includegraphics[width=0.44\textwidth]{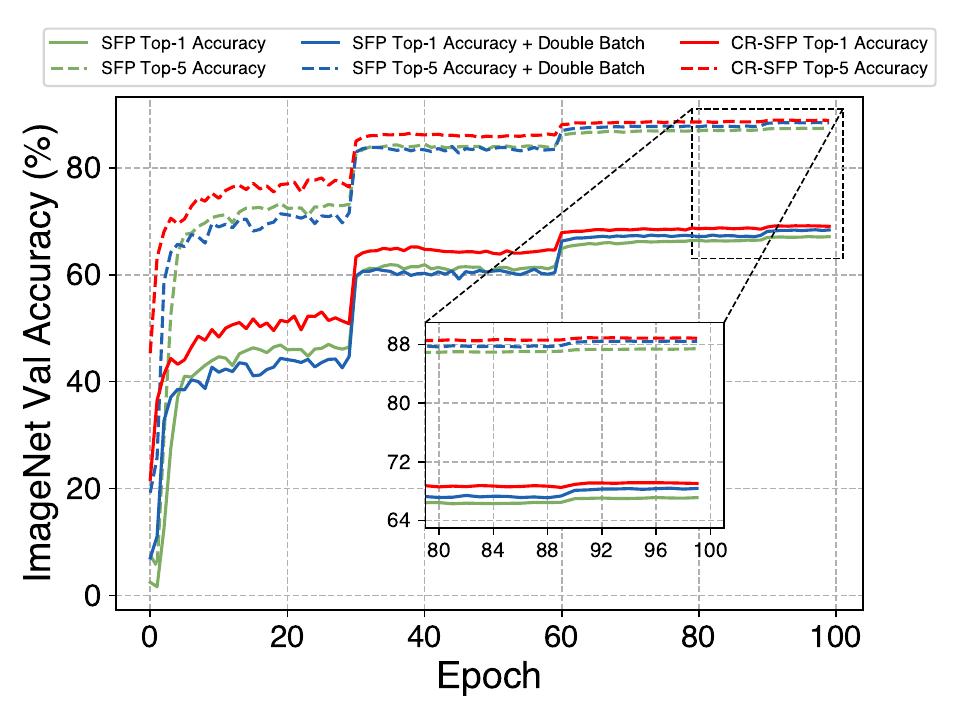}
    \vspace{-0.1in}
    \caption{ImageNet val accuracy of SFP, SFP with a doubled batch size and CR-SFP. Top-1 and Top-5 accuracy are reported.}
    \label{fig:convergence}
\end{figure}
\section{Conclusions}
SFP is an effective way to enhance the performance of pruning networks.
This paper focuses on solving the inconsistent representation problem of R-NN and P-NN during the training process.
Specifically, we propose learning consistent representation for soft filter pruning~(CR-SFP).
During the training procedure, CR-SFP optimizes R-NN and P-NN simultaneously in a parameter-shared manner,
Meanwhile, it improves their consistency by minimizing the posterior probability distribution from different data-distorted versions with bidirectional KL-divergence loss.
During the inference, we can export P-NN without introducing any additional computational costs.
CR-SFP is an effective and efficient method which not only inherits the advantage of SFP that can preserve model capacity but also improves the peformance of P-NN.
In the case of limited training epochs, extensive experiments have demonstrated that CR-SFP can achieve competitive or superior results compared to state-of-the-art methods.
Furthermore, transferring the consistent representation learning to other scenarios, such as binary neural networks (binary weights versus full precision weights), will be further studied in our future work.